\pdfoutput=1

\documentclass[11pt]{article}

\usepackage[]{acl}

\usepackage{times}
\usepackage{latexsym}

\usepackage{graphicx}
\usepackage{booktabs,multirow}
\usepackage{colortbl}
\usepackage{float}
\usepackage{amsmath}
\usepackage{stfloats}
\usepackage{amssymb}
\usepackage{pifont}
\newcommand{\orange}[1]{\textcolor[HTML]{EC6613}{#1}}
\newcommand{\ph}[1]{\phantom{#1}}
\newcommand{\tb}[1]{\textbf{#1}}

\newcommand{\cmark}{\ding{51}}%
\newcommand{\xmark}{\ding{55}}%

\usepackage[T1]{fontenc}

\usepackage[utf8]{inputenc}

\usepackage{microtype}

\usepackage{inconsolata}

\title{A Large Collection of Model-generated Contradictory Responses \\ for Consistency-aware Dialogue Systems}

\author{
Shiki\,Sato$^{1}$\hspace{1em}
Reina\,Akama$^{1,2}$\hspace{1em}
\textbf{Jun\,Suzuki}$^{1,2}$\hspace{1em}
\textbf{Kentaro\,Inui}$^{3,1,2}$\\[3pt]
$^{1}$Tohoku University\hspace{1em}
$^{2}$RIKEN\hspace{1em}
$^{3}$MBZUAI\hspace{1em}
\\\texttt{\{shiki.sato.d1,akama,jun.suzuki\}@tohoku.ac.jp}
\\\texttt{kentaro.inui@mbzuai.ac.ae}
}

\begin{document}
\maketitle
\begin{abstract}
Mitigating the generation of contradictory responses poses a substantial challenge in dialogue response generation.
The quality and quantity of available contradictory response data play a vital role in suppressing these contradictions, offering two significant benefits.
First, having access to large contradiction data enables a comprehensive examination of their characteristics.
Second, data-driven methods to mitigate contradictions may be enhanced with large-scale contradiction data for training.
Nevertheless, no attempt has been made to build an extensive collection of model-generated contradictory responses.
In this paper, we build a large dataset of response generation models' contradictions for the first time.
Then, we acquire valuable insights into the characteristics of model-generated contradictions through an extensive analysis of the collected responses.
Lastly, we also demonstrate how this dataset substantially enhances the performance of data-driven contradiction suppression methods.
\end{abstract}

\section{Introduction}
Recent large-scale neural response generation models (RGMs) have made significant progress~\cite{adiwardana:arxiv2020:meena,bao:acl2021:plato2,bao:acl2022:platoxl}.
However, they still struggle to generate semantically appropriate responses~\cite{roller:eacl2021:blenderbot,shuster:arxiv2022:bb3}.
Among various issues, contradictory responses%
    \footnote{Note that we focus on the contradictions against what is stated in the local context rather than those against the facts in the world. More details are described in Appendix~\ref{appendix:contradiction-types}.}
pose a particularly grave concern.
For example, in a conversation between speakers A and B, imagine that in response to speaker A's initial statement, \textit{I like tennis}, speaker B asks, \textit{How often do you play tennis?}
If speaker A then replies, \textit{I hardly ever play. I don't like tennis}, this response would be inconsistent with speaker A's initial statement.
Since these contradictions disrupt the dialogue flow and create a detrimental perception of the RGM's lack of comprehension of the dialogue content~\cite{nie:acl2020:i-like-fish,li:acl2022:mitigating}, effectively addressing them is crucial in developing RGMs to establish a trustworthy and symbiotic relationship with users.
Given this background, several works proposed approaches for mitigating contradictions (Section~\ref{sec:related-works}), but the problem remains unresolved and demands further improvements.

\begin{table}[t]
    \centering
    \footnotesize
    \tabcolsep 0.4mm
    \begin{tabular}[t]{rp{7.1cm}}
        \toprule
         & \multicolumn{1}{c}{Context} \\
        \midrule
        A1: & \tb{I hurt my toe doing ballet.} \orange{--- $u_r$} \\
        \rowcolor{gray!7}
        B1: & Oh I hope you get better. Does is hurt a lot? \\
        A2: & It hurts pretty bad, but it will heal. [...] \\
        \rowcolor{gray!7}
        B2: & [...] Do you do ballet practice often? \orange{--- ${u_q}$} \\
        \midrule
    \end{tabular}
    \begin{tabular}[t]{rp{5.74cm}}
         \multicolumn{2}{c}{A's next sample utterances responding to B2 ($u_q$)} \\
        \midrule
        \vspace*{0.1cm}
        A3 (RGM-1): & \tb{I don't do ballet myself,} I was just watching a performance. [...] \orange{--- [\xmark, \xmark, \xmark]} \\
        \vspace{-1ex}
        A3 (RGM-2): & \tb{I have never done ballet,} but I love the music. I listen to it all the time. \orange{--- [\xmark, \xmark, \xmark]} \\
        \multicolumn{2}{c}{\vdots} \\
        A3 (RGM-$n$): & Yes, I do ballet every day. [...] \orange{--- [\cmark, \cmark, \cmark]} \\
        \bottomrule
    \end{tabular}
    \caption{
    Example of annotated conversation in our dataset.
    The speakers are identified as A and B.
    \xmark{} and \cmark{} are labels provided by three human workers, indicating that a given A3 utterance (RGM-generated response) is contradictory or noncontradictory, respectively, with respect to the utterance specified by $u_r$.
    Contradictory segments are bolded for illustration.}
    \label{table:data-samples}
\end{table}

A significant obstacle hindering further progress in suppressing contradictions is the lack of a large-scale collection of RGM-generated contradictory responses.
This deficiency poses two challenges.

First, the limited availability of such data impedes studies for understanding the nature of RGM-generated contradictions.
For instance, investigating the correlation between the presence of a certain feature (e.g., a specific dialogue act label) in a dialogue context and the occurrence of an RGM contradiction can aid in developing more effective strategies for mitigating contradictions.
Regrettably, the available resources are limited to \citet{nie:acl2020:i-like-fish}'s small collection (a few hundred) of RGM-generated contradictory instances intended as test data, making it insufficient for such investigation.

Second, the efficacy of data-driven contradiction suppression may be limited by a scarcity of training data.
As evidenced in various NLP tasks~\cite{leite:aacl2020:toxic,mosbach:iclr2021:stability}, the performance of data-driven systems is dependent on the volume of available data.
Therefore, the efficacy of data-driven contradiction suppression could be improved with access to large-scale RGM contradiction data.
Although data-driven approaches have been discussed in previous studies (Section~\ref{subsec:existing-methods}), they were based on alternative resources such as automatically synthesized contradictions or human-written contradictions~\cite{nie:acl2020:i-like-fish,li:acl2022:mitigating}.
However, contradictions generated by automatic synthesis or manually are different in characteristics from those actually generated by RGMs (Section~\ref{subsec:analyze-response-itself}).
If one tries to handle RGMs' contradictions with models trained on alternative resources, the potential of data-driven methods may not be fully realized because of the discrepancy between the training data and the practical inference targets, as demonstrated in Section~\ref{sec:experiments}.

In this paper, we demonstrate the effectiveness of having a vast repository of RGM-generated contradictory responses in tackling RGM contradictions.
To begin with, we build a large-scale dataset comprising $10$K contradictory and $17$K noncontradictory responses generated by various high-performance RGMs.
The consistency of each response is judged by three human annotators, as illustrated in Table~\ref{table:data-samples}.
To our knowledge, this is the first work to construct a dataset containing more than $1$K contradictory RGM responses with human annotations.
We then analyze our collection from various angles, yielding valuable insights into RGM contradictions.
We also demonstrate that a contradiction detector trained on human-written contradiction data exhibits limited accuracy in identifying RGM contradictions, and training on our dataset improves this situation.
Our dataset will be made publicly available.


\section{Related studies}
\label{sec:related-works}

\subsection{Major methods to handle contradictions}
\label{subsec:existing-methods}
The mainstream approaches of prior studies to mitigate contradictions have been data-driven.
\citet{welleck:acl2019:dialog-NLI} developed a dialogue-domain natural language inference dataset by applying a rule-based method to transform an existing dialogue corpus.
They employed this dataset to train a contradiction detector that automatically identifies contradictions within pairs of dialogue domain sentences.
\citet{nie:acl2020:i-like-fish} gathered and used $15,\!605$ contradictory and $15,\!605$ noncontradictory human-written dialogue utterances to train a contradiction detector.
They also collected $382$ RGM-generated contradictory responses as test data to evaluate detectors.
Meanwhile, \citet{li:acl2020:dontsaythat} and \citet{li:acl2022:mitigating} updated RGMs using a loss function that reduces the likelihood of generating inconsistent responses.
We aim to collect RGM-generated contradictions to provide valuable resources for these data-driven methods.
To our knowledge, this is the first work to gather a substantial volume of contradictory responses from RGMs.

\subsection{Effective inputs to collect contradictions}
\label{subsec:relatedworks-context}
Previous studies demonstrated that RGMs tend to generate contradictions when they repeat previously stated facts or opinions~\cite{nie:acl2020:i-like-fish, li:acl2021:addressing}.
Nevertheless, posing questions that prompt dialogue partners to repeat previously stated information can be uncommon in natural dialogues.
On the other hand, we aim to collect RGM contradictions by identifying contradictions in RGM responses to follow-up questions.
Follow-up questions seek additional information related to the information previously stated by the dialogue partner.
These types of questions commonly arise during dialogues.
Follow-up questions are similar to the abovementioned queries (i.e., questions requesting repetitions of previously mentioned facts or opinions) as they both seek information related to the previously stated content.
With this similarity, we hypothesize that follow-up questions also tend to induce RGM contradictions.


\section{Dataset construction}
\label{sec:data-collection}
This paper showcases the importance of employing extensive datasets containing RGM-generated contradictory and noncontradictory responses to mitigate RGM contradictions effectively.
As stated earlier, large-scale data are currently lacking.
To address this issue, we first performed an extensive collection of RGM-generated instances.
This section outlines the methodology used to build our dataset, followed by the detailed settings and the data collection results for this study.

\subsection{Method}

\begin{figure}[t]
\begin{center}
\includegraphics[width=\columnwidth]{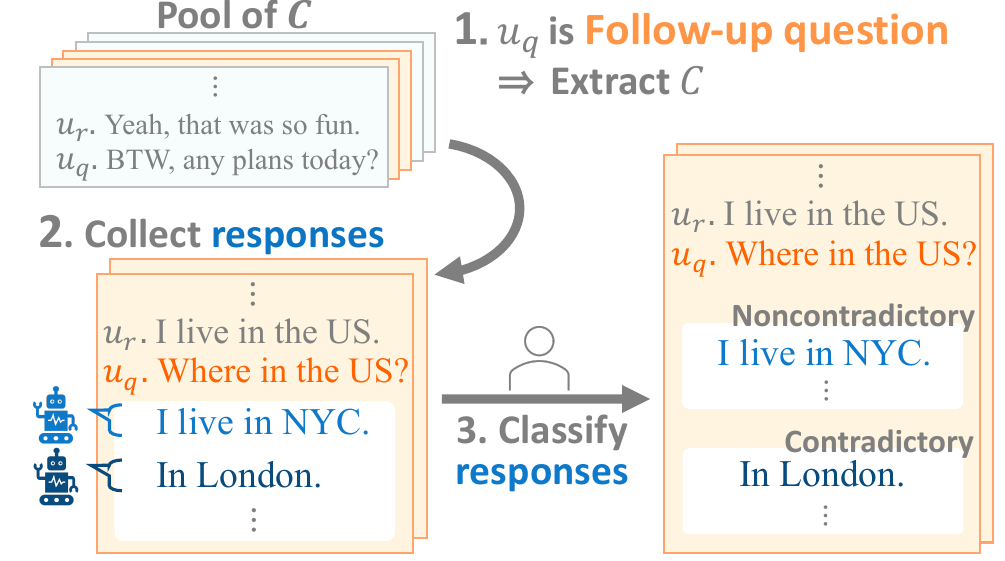}
\end{center}
\caption{Overview of our data collection process.}
\label{fig:collection-method}
\end{figure}

Figure~\ref{fig:collection-method} illustrates our data collection process.
We first prepare dialogue contexts as input and then collect their RGM responses.
The collected responses are classified into contradictory or noncontradictory groups according to their contexts.
This process is based on that used by \citet{nie:acl2020:i-like-fish}, with the differences being the approach to the context preparation and the focus on RGM responses instead of human-written ones.

\subsubsection{Dialogue context preparation}
\label{subsubsec:fqness}

Contradictory responses are inconsistent with contexts; hence, their occurrences depend on their contexts.
For instance, it is improbable that a contradiction will occur in a context where only greetings are exchanged.
Based on previous insights (Section~\ref{subsec:relatedworks-context}) and our preliminary analysis (Appendix~\ref{appendix:whyfq}), we gathered follow-up questions (FQs) as the prime contexts for eliciting contradictions.

RGMs do not generate contradictions exclusively to FQs; hence, addressing all contradiction types solely by examining the contradictions to FQs is impractical.
Nevertheless, we believe that refining contradiction suppression techniques using contradictory responses to these representative inputs can establish the groundwork for attempts to mitigate contradictions in a broader input range.
In fact, the experimental results in Section~\ref{subsec:result} show that a contradiction detector trained on our collection effectively identified the contradictions in responses to non-FQ contexts.

\begin{figure}[t]
\begin{center}
\includegraphics[width=\columnwidth]{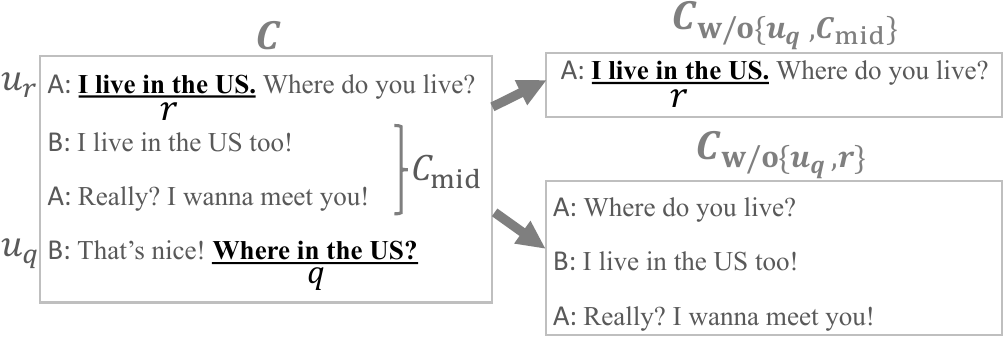}
\end{center}
\caption{Example of $C_{\mathrm{w/o}\{u_q, C_{\mathrm{mid}}\}}$ and $C_{\mathrm{w/o}\{u_q, r\}}$.}
\label{fig:fqness}
\end{figure}

\paragraph{Idea for collecting FQ.}
In Figure~\ref{fig:fqness}, $C$ refers to a dialogue context concluding with an utterance%
    \footnote{\,``Utterance'' in this study refers to all sentences in a turn.}
$u_q$ that contains a question $q$.
We use $u_r \in C$ to represent the utterance that precedes $u_q$ by $d_{u_r}$ utterances.
Note that we only consider scenarios wherein the $u_r$ and $u_q$ speakers are distinct individuals.
When $q$ is a question that refers to a specific segment $r$ in $u_r$,%
    \footnote{Let $q$ represent all interrogative sentences within an utterance, while $r$ encompasses all other sentences in the same utterance. Any sentences terminated with a question mark were identified as interrogative. The segmentation of an utterance into individual sentences was accomplished utilizing spaCy (\texttt{en\_core\_web\_sm})~\cite{spacy2}.}
as illustrated on the left side of Figure~\ref{fig:fqness}, we regard $q$ as an FQ for $r$.
Throughout this paper, the segment $r$ to which $q$ refers will be termed ``the referent of $q$.''
To determine whether $r$ is the referent of $q$, i.e., whether $q$ is an FQ for $r$, we must check whether $r$ and $q$ are relevant.
Fortunately, recent neural RGMs can generate highly relevant responses to contexts~\cite{zhang:acl2020demo:dialogpt,adiwardana:arxiv2020:meena}; hence, these RGMs are expected to effectively capture the relevance between utterances.
In other words, if a large-scale neural RGM deduces the strong relevance between $r$ and $q$, we can reasonably consider $q$ as an FQ for $r$.
Here, we introduce a new automatic metric that uses a neural RGM to assess the relevance between $q$ and $r$.

\paragraph{Method for collecting FQ.}
As Figure~\ref{fig:fqness} shows, $C_{\mathrm{w/o}\{u_q, C_{\mathrm{mid}}\}}$ refers to $C$, excluding $u_q$ and the intervening utterances $C_{\mathrm{mid}}$ between $u_r$ and $u_q$.
Similarly, $C_{\mathrm{w/o}\{u_q, r\}}$ represents $C$ with both $u_q$ and $r$ removed. 
If $q$ is an FQ for $r$, it is improbable for $P(u_q|C_{\mathrm{w/o}\{u_q, C_{\mathrm{mid}}\}})$ to exhibit a decrease compared to $u_q$'s original conditional probability.
Moreover, $P(u_q|C_{\mathrm{w/o}\{u_q, r\}})$ is likely lower than $u_q$'s original conditional probability.
Consequently, the following value, \tb{FQness}, is deemed high when $q$ is an FQ for $r$:
\begin{equation*}
    P(u_q|C_{\mathrm{w/o}\{u_q, C_{\mathrm{mid}}\}})/P(u_q|C_{\mathrm{w/o}\{u_q, r\}}).
\end{equation*}
Here, we compute the probabilities using an RGM.\footnote{We employed Blenderbot~\cite{roller:eacl2021:blenderbot} implemented in ParlAI~\cite{miller:emnlp2017demo:ParlAI}, a well-known high-performance RGM.}
We collect FQs by selecting samples with the highest FQness scores from a pool of $C$.
Appendix~\ref{appendix:validate-fq} illustrates the preliminary experiment to assess the effectiveness of picking contexts based on FQness for efficiently gathering RGM contradictions.

\subsubsection{RGM response collection}
For every gathered $C$ with a high FQness, multiple RGMs generate responses to gather diverse contradictions from various RGMs efficiently.

\subsubsection{RGM response annotation}
We assign three human workers to assess each generated response and categorize it into two groups, contradictory and noncontradictory, according to their preceding referent $r$ in $u_r$.
If at least two workers determine the presence of contradictions in a response, this response is labeled contradictory.
If all workers agree that a response is consistent, this response is labeled as noncontradictory.

\subsection{Construction settings}
\subsubsection{Settings of dialogue context preparation}
\label{subsubsec:settings-inputs}

A pool of $C$ is formed by extracting $d_{u_r}$ or more consecutive utterances from dialogue corpora, ensuring that the final utterance includes questions.
From this pool, we gather those with the highest FQness scores.
For this study, we gathered $C$ from the Multi-session Chat (MSC) dataset~\cite{xu:acl2022:goldfish}.
This dataset exhibits distinctive features that make it an ideal source for collecting $C$: ($1$) low noise (e.g., few misspellings), and ($2$) realistic dialogues between acquaintances, wherein speakers engage in in-depth discussions on various topics.

Since the annotation cost extremely increases as the value of $d_{u_r}$ increases,%
\footnote{As written in Section~\ref{subsubsec:fqness}, $d_{u_r}$ refers to the distance between $u_r$ and $u_q$.
When the value of $d_{u_r}$ is large, $q$ denotes the FQ associated with the earlier utterance in the context.}
the highest value assigned to $d_{u_r}$ was $5$ for this study, and we intensively gathered FQs with $d_{u_r}=1$, $3$ to prepare for collecting contradictory responses.
From the set of approximately $59$K $C$ in the MSC dataset, we extracted $3,\!250$, $1,\!000$, and $100$ samples for $d_{u_r}=1$, $3$, and $5$, respectively, based on the FQness scores.

\subsubsection{RGM response collection}
\label{subsubsec:settings-responses}

We employed eight high-performance RGMs: Plato-2 (P2), Plato-XL (PX)~\cite{bao:acl2021:plato2,bao:acl2022:platoxl}, Blenderbot1-3B (B1), Blenderbot2-3B (B2)~\cite{xu:acl2022:goldfish, acl2022:komeili:internet}, Blenderbot3-3B (B3)~\cite{shuster:arxiv2022:bb3}, Blenderbot3-30B (BL), Opt-66B (O6)~\cite{zhang:arxiv2022:opt}, and ChatGPT (CG).\footnote{\url{https://openai.com/chatgpt}.}
Each RGM generated a response to an input, resulting in eight responses for each $C$.
Appendix~\ref{appendix:generation-settings} presents detailed settings.

\subsubsection{RGM response annotation}
\label{subsubsec:settings-annotator}
We employed Amazon Mechanical Turk\footnote{\url{www.mturk.com}.} to recruit workers.
We ensured the creation of a high-quality, cost-effective data set by carefully selecting highly skilled workers.
The selection procedure is shown in Appendix~\ref{appendix:worker-selection}.
During the data collection phase, we published tasks to workers that required classifying $40$ responses for five $C$.

\subsection{Construction results}

\begin{table}[t]
    \small
    \centering
    \tabcolsep 12.8pt
    \begin{tabular}{ccc}
        \toprule
         Val. of $d_{u_r}$ & \# of C                & \# of N               \\
        \midrule
         1           & \ph{0}8108 \,(\ph{}2703) & \ph{}12471 \,(\ph{}2920) \\
         3           & \ph{0}2175 \,(\ph{0}739) & \ph{0}4378 \,(\ph{0}953) \\
         5           & \ph{00}220 \,(\ph{00}74) & \ph{00}422 \,(\ph{00}94) \\
        \midrule
         Total       & \ph{}10503 \,(\ph{}3516) & \ph{}17271 \,(\ph{}3967) \\
        \bottomrule
    \end{tabular}
    \caption{Summary of our dataset. ``\# of C'' and ``\# of N'' denote the numbers of the collected contradictory and noncontradictory responses, respectively. The values in parentheses refer to the number of unique contexts. More detailed statistics are shown in Appendix~\ref{appendix:breakdown}.}
    \label{table:dataset-statistics}
\end{table}

Table~\ref{table:dataset-statistics} reports several dataset statistics.
Throughout the annotation process, the groups of three workers achieved average Fleiss' kappa values of $0.405$, $0.465$, and $0.408$ for $d_{u_r}=1$, $3$, and $5$, respectively.
Given the intricacies involved in identifying contradictions, the substantial level of consensus signified the successful creation of a high-quality dataset.
Table~\ref{table:data-samples} provides examples of our dataset.
Each sample comprises a dialogue context $C$ containing $u_r$ and $u_q$, and an RGM response contradictory or noncontradictory to $r$ in $u_r$ with the labels assigned by the three workers.


\section{Dataset analysis}
\label{sec:analyze-contradiction}
Analyzing the characteristics of RGM contradictions is crucial in devising innovative approaches to mitigate contradictions.
The examinations performed by prior studies have been limited by the lack of extensive data, a gap that our dataset effectively fills.
Sections~\ref{subsec:analyze-response-itself} and \ref{subsec:analyze-context} delineate the results of analyzing these characteristics using our data set.
Section~\ref{subsec:analyze-response-itself} delves into the intrinsic features of the generated responses, while Section~\ref{subsec:analyze-context} presents an examination centered around the dialogue contexts that trigger RGM contradictions.

\subsection{Analysis of RGM-generated responses}
\label{subsec:analyze-response-itself}
Our analysis identified two types of characteristic contradictions in the RGM responses: contradictions arising from intra-utterance inconsistencies and those related to ambiguous expressions.

\begin{table}[t]
    \centering
    \footnotesize
    \tabcolsep 0.5mm
    \begin{tabular}[t]{cp{7.15cm}}
        \toprule
         & \multicolumn{1}{c}{Context} \\
        \midrule
        A: & I made plans to travel to a new place next month. \\
        \rowcolor{gray!7}
        B: & What attracted you to this new place? Where is it? \\
        \midrule
         & \multicolumn{1}{c}{RGM responses on speaker A's side} \\
        \midrule
    \end{tabular}
    \begin{tabular}[t]{cp{7cm}}
        O6: & I've been to this place before and I really liked it. It's in a country I've never been to before.\\
        \bottomrule
    \end{tabular}
    \caption{An example contradictory response by Opt-66B with an intra-utterance inconsistency.}
    \label{table:intra}
\end{table}

\subsubsection{Intra-utterance inconsistencies}
\label{subsubsec:difference-manner}
A qualitative analysis of the RGM contradictions in our dataset revealed that one distinctive way contradictions occur is through intra-utterance inconsistency.
Table~\ref{table:intra} provides an example of this type of contradiction.
In this conversation, Opt-66B generated inconsistent information within a single utterance, saying, \textit{I've been to this place} while stating, \textit{It's in a country I've never been to.}
When conflicting information coexists within a statement, it becomes highly probable that at least one of them contradicts the context.
Instances of contradictions stemming from intra-utterance inconsistencies are occasionally observed across multiple RGMs.
To delve deeper into this phenomenon, we counted the inconsistencies among the $50$ randomly selected contradictory responses in our dataset for each of the eight RGMs.
Our findings indicated that seven RGMs generated at least $4$ ($8\%$) contradictory responses featuring an intra-utterance inconsistency.
Conversely, none of the $200$ human-written contradictory responses randomly sampled from the DECODE dataset exhibited an intra-utterance inconsistency.
See Appendix~\ref{appendix:intra} for detailed results.
These results suggest contradictory responses featuring intra-utterance inconsistencies are particularly frequent in RGM responses.

\subsubsection{Ambiguous expression}
\label{subsubsec:difference-ambiguity}

\begin{table}[t]
    \centering
    \footnotesize
    \tabcolsep 0.5mm
    \begin{tabular}[t]{cp{7.15cm}}
        \toprule
         & \multicolumn{1}{c}{Context} \\
        \midrule
        A: & I had a promising interview today! \\
        \rowcolor{gray!7}
        B: & Oh excellent! How did it go, what made it so excellent? \\
        \midrule
         & \multicolumn{1}{c}{RGM responses on speaker A's side} \\
        \midrule
    \end{tabular}
    \begin{tabular}[t]{cp{7cm}}
        P2: & i think i did well because they called me back to set up \tb{an interview}.\\
        \bottomrule
    \end{tabular}
    \caption{Example of Plato-2's contradictory responses with ambiguity. The determination of whether or not a contradiction exists hinges upon the interpretation assigned to the bolded term ``interview,'' particularly if it is construed to differ from the preceding interview.}
    \label{table:ambiguity}
\end{table}

We observed a notable distinction in the human annotation tendency on the existence of contradictions between the set of human-written responses in DECODE and our compilation of RGM-generated responses.
Both our study and \citet{nie:acl2020:i-like-fish} employed a similar approach in selecting the human workers who identified contradictions during the data creation process (Section~\ref{subsubsec:settings-annotator}).
However, within the subset of instances where at least one worker detected the contradictions, a significant gap was observed in the proportions where the other two workers also concurred on the existence of contradictions.
This proportion was $78.4\%$ for the human-written responses and $30.4\%$ for the RGM-generated ones.
This dissimilarity could have stemmed from the RGM's propensity to generate ambiguous expressions concerning consistency, as demonstrated in Table~\ref{table:ambiguity}.
Such responses appeared to result in differing judgments regarding the presence of contradictions, depending on how individual workers interpreted them.
Suppressing these contradictions is crucial, even if some workers may miss the inconsistencies, because they, once perceived as contradictory by actual users, can significantly detriment the quality of dialogues.

\subsection{Analysis of dialogue contexts}
\label{subsec:analyze-context}
If specific dialogue contexts induce contradictory responses from various RGMs, identifying their contributing characteristics may become crucial in developing more effective contradiction mitigation techniques.
Our dataset is suitable for this investigation because it contains a lot of $C$ for which diverse RGMs generate responses.
We conducted an examination to identify features of $C$ that induce contradictions from a lot of RGMs, employing statistical tests on our dataset.%
\footnote{Initially, we categorized each of $C$ into two sets based on the presence or absence of a certain feature, such as whether the utterance $u_q$ contains the word ``how.''
Subsequently, for each of these two sets, we computed the average number of contradictory responses elicited by one $C$ from the eight RGMs.
We regarded a feature of $C$ as the one inducing many RGM contradictions if a statistically significant difference in the average number between the two sets was identified with a one-tailed t-test at a $1$\% significance level.}
It is noteworthy that this type of statistical analysis becomes feasible due to the creation of a large collection of RGM-generated contradictions, such as our dataset.
Our analysis focused on dialogue act labels and lexical attributes, which are highly interpretable and seem particularly well-suited as focal points of the first analysis.

\begin{table}[t]
    \centering
    \footnotesize
    \tabcolsep 0.5mm
    \begin{tabular}[t]{cp{7cm}}
        \toprule
         & \multicolumn{1}{c}{Context} \\
        \midrule
        A: & Have you taken any new pictures? \\
        \rowcolor{gray!7}
        B: & I managed to get out at the weekend and get loads of shots in the snow we had. [. . .] \\
        A: & Oh wow you had snow!? \tb{We just had rain all weekend :)} [. . .] Did you have a nice chilled weekend? [. . .] \\
    \end{tabular}
    \begin{tabular}[t]{cp{7cm}}
        \midrule
         & \multicolumn{1}{c}{RGM responses on speaker B's side} \\
        \midrule
        \rowcolor{gray!7}
        P2: & it was a good weekend here, \tb{we got to enjoy the cold rain}! \\
        \bottomrule
    \end{tabular}
    \caption{Example of Plato-2's contradictory responses containing a partner's bolded statement.}
    \label{table:statement}
\end{table}

\paragraph{Analysis results of dialogue acts.}
When we assigned SWBD-DAMSL dialogue act labels~\cite{jurafsky:97:switchboard} to $u_q$ in our dataset,%
    \footnote{See Appendix~\ref{appendix:dialogue-act} for detailed labeling settings.}
we observed a notable trend, that is, $u_q$ categorized as `Declarative Yes-No-Questions' or `Statement-non-opinion' were more prone to triggering contradictions.
Among the $193$ assigned instances for the former label, the average count of the contradictory responses from the eight RGMs per $C$ was $2.77$ (i.e., $2.77$ contradictory responses / $8$ generated responses $=35\%$).
The average for the $4084$ unassigned instances was lower at $2.41$ ($30\%$).
This phenomenon could have arisen from a deficiency in the RGM's ability to generate appropriate responses while being cognizant that a repetition of previous information is being solicited.
Focusing on $2,\!118$ assigned instances for the latter label indicated a higher average of $2.49$ ($31\%$) contradictory responses compared to an average of $2.36$ ($30\%$) for the $2,\!159$ unassigned ones.
This disparity could arise from the RGMs' inability to differentiate between the dialogue partners' statements and their own utterances in dialogue contexts.
Hence, RGMs may generate responses incorporating the partners' information as if it were their own, even if it is inconsistent with their past statements, as exemplified in Table~\ref{table:statement}.

\paragraph{Analysis results of lexical features.}
The $u_q$ containing the interrogative term ``how'' can provoke contradictions.
More precisely, the mean count of the contradictory responses within the $764$ applicable contexts stood at $2.60$ ($33\%$), while that in the $3,\!513$ inapplicable contexts was $2.39$ ($30\%$).
Answering ``How questions'' while upholding consistency with the context poses a challenge for the current RGMs.


\section{Experiments}
\label{sec:experiments}
This section presents compelling evidence to support the hypothesis that employing the RGM-generated contradiction collection as a training resource yields notable enhancements in the effectiveness of data-driven contradiction suppression methods.
As a case in point, we focused on developing a contradiction detector that automatically classifies whether or not a given utterance pair is contradictory by training it on our dataset.
Contradiction detectors are commonly employed in post-processing tasks that filter out RGMs' contradictory response candidates~\cite{welleck:acl2019:dialog-NLI,nie:acl2020:i-like-fish} and automatic evaluations of RGMs' contradiction frequencies~\cite{li:acl2021:addressing}, effectively playing a crucial role in mitigating contradictions.

Existing detectors have been developed by employing automatically synthesized or human-written contradictions as substituting training resources for RGM contradictions.
We hypothesize that their performance can be enhanced by utilizing RGM contradiction data for their training.
Our experiments validate the potency of our dataset by assessing the contradiction detection performance of a detector trained on our dataset against that of a detector trained with human-written contradictions.

\subsection{Settings}
\label{subsec:settings}
\paragraph{Inputs and outputs.}
Like the utterance-based detectors of \citet{nie:acl2020:i-like-fish}, given a dialogue response and the corresponding preceding utterance $u_r$, a detector yields a binary classification result indicating whether the response contradicts $u_r$.

\paragraph{Contradiction detectors.}
We conducted a performance analysis of a detector that underwent training on our dataset, juxtaposed with a detector fashioned similarly to the state-of-the-art detector devised by \citet{nie:acl2020:i-like-fish}.
Their detector was developed by fine-tuning RoBERTa~\cite{liu:arxiv2019:roberta} on the DECODE dataset specifically for the binary classification tasks requiring the prediction of consistency within a pair of given utterances.
Following their settings, we developed a Contradiction Detector by fine-tuning RoBERTa on our dataset, denoted as $\mathrm{CD_{OUR}}$.
Similarly, we constructed a rival detector, $\mathrm{CD_{DEC}}$, using an equivalent number of instances from the DECODE dataset as $\mathrm{CD_{OUR}}$.

\paragraph{Training data for $\mathbf{CD_{OUR}}$.}
Our dataset contains both contradictory and noncontradictory responses from eight RGMs.
Our experiment performed a cross-validation test by selecting one RGM (i.e., target RGM) and using its responses as the test data.
The samples excluding the target RGM's responses were used for training.
We realized a comprehensive assessment of the detectors' performance by conducting the evaluation process eight times, varying the target RGMs each time.
When we selected B2 as the target model, the number of training data samples was minimized to $8,\!023$ contradictory and $8,\!023$ noncontradictory responses; we reduced the number of training data samples to align with this number when we specified one of the other RGMs as a target RGM.
Appendix~\ref{appendix:training} presents the training details.

\paragraph{Training data for $\mathbf{CD_{DEC}}$.}
We randomly selected $8023$ contradictory and $8,\!023$ noncontradictory human-written responses from the DECODE dataset.
Other settings are the same as  $\mathrm{CD_{OUR}}$.

\begin{table*}
    \centering
    \small
    \tabcolsep 3.18mm
    \begin{tabular}{lcclccccccccc}
        \cmidrule[0.9pt]{1-2} \cmidrule[0.9pt]{4-12}
            Detector            & By-Human  &           & Detector &P2        & PX        & B1
                                & B2        & B3        & BL        & O6        & CG        \\
        \cmidrule{1-2} \cmidrule{4-12}
            $\mathrm{CD_{DEC}}$ &\tb{.952} &           & $\mathrm{CD_{DEC}}$&.600       &.575       &.615
                                &.540       &.655       &.555       &.565       &.650       \\
            $\mathrm{CD_{OUR}}$ & .842      &           & $\mathrm{CD_{OUR}}$&\tb{.800} &\tb{.735} &.\tb{715}
                                &\tb{.750} &\tb{.765} &\tb{.745} &\tb{.690} &\tb{.790} \\
        \cmidrule[0.9pt]{1-2} \cmidrule[0.9pt]{4-12}
        \multicolumn{2}{c}{(a) Human-written test set.} & &
        \multicolumn{9}{c}{(b) In-domain test sets. Scores for each target RGM are presented.} \\
    \end{tabular}
    \tabcolsep 2.95mm
    \begin{tabular}{lccccccccc}
        \toprule
                                & \footnotesize{Test set from Nie+'21}  &
                                & \multicolumn{7}{c}{\footnotesize{Test sets from Topical-Chat\,/\,DailyDialog}} \\
        \cline{2-2} \cline{4-10} \\[-1.8ex]
            Detector            & Human-Bot &           & P2                 & PX                 & B1
                                & B2                    & B3                 & BL                 & O6 \\
        \midrule
            $\mathrm{CD_{DEC}}$ & .749      &           &.55/.52             &.58/.60             &.61/.55
                                &.60/.59                &.68/.61             &.67/.55             &.59/.53 \\
            $\mathrm{CD_{OUR}}$ &\tb{.787} &           &\tb{.77}/\tb{.77} &\tb{.75}/\tb{.71} &\tb{.74}/\tb{.68}
                                &\tb{.70}/\tb{.72}    &\tb{.73}/\tb{.76} &\tb{.82}/\tb{.64} &\tb{.81}/\tb{.75} \\
        \bottomrule
        \multicolumn{10}{c}{(c) Out-of-domain test sets. For the Topical-Chat and DailyDialog test sets, scores for each target RGM are presented.} \\
    \end{tabular}
    \caption{Accuracy of the detectors for (a) human-written, (b) in-domain, and (c) out-of-domain test sets. $\mathrm{CD_{OUR}}$'s score for By-Human is the median of $\{0.819, 0.827, 0.838, 0.840, 0.843, 0.847, 0.859, 0.871\}$ since we trained the eight $\mathrm{CD_{OUR}}$ detectors as explained in Section~\ref{subsec:settings}. $\mathrm{CD_{OUR}}$'s score for Human-Bot refers to the accuracy of the detector trained without B1's responses because Human-Bot contains B1's responses.}
    \label{table:detector-accuracy-indomain}
\end{table*}

\paragraph{In-domain test sets.}
We randomly selected $100$ contradictory and $100$ noncontradictory responses of the target RGM responses from our dataset as test samples.
Note that a training set might also contain responses of non-target RGMs that share the same contexts as these $200$ test samples.
We excluded these samples from the training set to ensure a fair evaluation of the detectors' ability to identify contradictions from unknown RGMs for unfamiliar contexts.

\paragraph{Out-of-domain test sets.}
The above RGM-generated test sets were derived from the corpus used to develop the training set for $\mathrm{CD_{OUR}}$.
Furthermore, these sets exclusively comprise responses to FQs.
To assess the detector's effectiveness in identifying contradictions in RGM responses to non-FQ contexts from unfamiliar dialogue corpora, we prepared two out-of-domain test sets.
One set originated from the Topical-Chat dataset~\cite{gopalakrishnan:interspeech2019:topical}, and the other from the DailyDialog dataset~\cite{li:ijcnlp2017:dailydialog}.
Each of these sets comprises seven subsets, each containing $50$ contradictory and $50$ noncontradictory responses from P2, PX, B1, B2, B3, BL, or O6.%
    \footnote{CG was omitted from the test set construction due to cost considerations, as CG's contradiction frequency was low (Appendix~\ref{appendix:breakdown}).}
The contexts of these sets were randomly selected from all contexts concluding with utterances containing questions not limited to FQs, in the corpora.%
    \footnote{Considering that non-question contexts may allow contextually irrelevant replies, such as prompting changes in the topic, we anticipate a lower occurrence of contradictions. Our focus on responses only to questions is in accordance with cost considerations.}
Appendix~\ref{appendix:testset} details the construction process.
The subsets from these two test sets were used in a manner resembling a cross-validation test, akin to how the subsets of the in-domain test set were employed, except that even the subsets of non-target RGMs' responses were excluded from the training set to prevent detectors from being trained on the same domain data.
In addition, we employed \citet{nie:acl2020:i-like-fish}'s Human-Bot dataset, which possesses $382$ contradictory and $382$ noncontradictory RGM responses in human-bot dialogues.

\paragraph{Human-written test set.}
We utilized \citet{nie:acl2020:i-like-fish}'s By-Human test set comprising $2108$ contradictory and $2,\!108$ noncontradictory human-written responses.
This allowed us to verify that $\mathrm{CD_{DEC}}$ was reasonably well-trained in our settings, although detecting human-written contradictions falls beyond the scope of our study.

\subsection{Results}
\label{subsec:result}
Table~\ref{table:detector-accuracy-indomain} (a), (b), and (c) display the accuracy of the contradiction detectors for the human-written, in-domain, and out-of-domain test sets, respectively.

\paragraph{(a) Human-written test set.}
$\mathrm{CD_{DEC}}$ obtained a high accuracy of $0.952$ on the By-Human test set, confirming that $\mathrm{CD_{DEC}}$ was properly trained.

\paragraph{(b) In-domain test sets.}
However, $\mathrm{CD_{DEC}}$ achieved low accuracy for the subsets of our RGM-generated dataset.
Particularly, it had an accuracy of only $0.540$ when B2 was the target RGM, which is problematic in practical applications.
In contrast, $\mathrm{CD_{OUR}}$ gained higher accuracy on our RGM-generated test sets.
The training process for $\mathrm{CD_{OUR}}$ excluded any contradiction data from the target RGMs and samples that shared the same dialogue contexts as the test data, thereby effectively detecting contradictions from unknown RGMs when confronted with unfamiliar contexts.

\paragraph{(c) Out-of-domain test sets.}
For all three test sets, the performance of $\mathrm{CD_{OUR}}$ significantly outperformed $\mathrm{CD_{DEC}}$.
These results emphasize that detectors trained on our dataset can effectively detect contradictions in RGM responses for contexts that are out-of-domain and non-FQ.

\paragraph{}
Note that it has been confirmed that $\mathrm{CD_{OUR}}$ exhibited superior performance even when the entirety of DECODE's samples was employed for training $\mathrm{CD_{DEC}}$, although the above experiments employed only approximately half of DECODE's samples during $\mathrm{CD_{DEC}}$'s training.
Furthermore, we have verified that $\mathrm{CD_{OUR}}$ outperforms all seven detectors constructed similarly to the seven baseline detectors employed in \citet{nie:acl2020:i-like-fish}'s experiments.
See Appendix~\ref{appendix:additional-experiments} for detailed results.

\subsection{Analysis: Performance improvement in detecting RGM-specific contradictions}
\label{subsec:analyze-result}
The above results exhibited that training detectors with the RGM contradictions led to a noticeable enhancement in the detectors' capability to identify RGM contradictions.
We hypothesized that this outcome could be attributed, at least in part, to the training on the RGM-generated instances, which facilitated the acquisition of identifying features typical of RGM contradictions, encompassing those expounded upon in Section~\ref{subsec:analyze-response-itself}.
We investigated the hypothesis' validity by taking the contradiction type mentioned in Section~\ref{subsubsec:difference-ambiguity}, contradiction with ambiguous expression, as an example.

\paragraph{Method.}
Our experiments revealed that $1,\!377$ RGM contradictory responses from our in-domain test sets and the validation sets used when training $\mathrm{CD_{OUR}}$ (see Appendix~\ref{appendix:training} for details) were missed by $\mathrm{CD_{DEC}}$ but successfully flagged by $\mathrm{CD_{OUR}}$.
Plausibly, some of these instances may exhibit certain features inherent to the RGM contradictions, which the training with RGM-generated data facilitated $\mathrm{CD_{OUR}}$ to recognize.
Therefore, we investigated if the $1,\!377$ contradictory responses encompass the distinguishing characteristic, i.e., ambiguous expression.

\paragraph{Results.}
All contradictory samples in our dataset were deemed contradictory by either two or three workers.
The instances judged contradictory by only two workers may encompass ambiguities regarding consistency.
Within those above $1,\!375$ contradictory responses, the proportion of the samples classified as contradictory by only two workers amounted to $51.3\%$.
Conversely, among the $4,\!378$ contradictory responses from our validation and test sets that both $\mathrm{CD_{DEC}}$ and $\mathrm{CD_{OUR}}$ successfully identified, only $43.3\%$ of the samples were determined contradictory by two workers.
This proportion gap exhibited statistical significance at the $1\%$ significance level in the chi-square test, underscoring that training on RGM-generated data enhanced the detector's capacity to recognize the contradictions characterized by ambiguity.


\section{Conclusion}
No attempt has been made to build an extensive collection of RGM-generated contradictory responses, which is problematic in two aspects: the scarcity of data for analysis and training.

In this paper, we built a large collection of contradictions generated by various RGMs for the first time.
We comprehensively analyzed our collection, producing valuable insights into the RGM contradictions, which we believe are crucial for effective contradiction suppression.
We also demonstrated that a contradiction detector trained on our dataset could identify RGM contradictions effectively.

Future challenges include applying the collected dataset to other data-driven methods and collecting data with a broader context variety than FQs.

\clearpage

\clearpage
\section*{Ethical concerns}
This study uses existing datasets, the MSC~\cite{xu:acl2022:goldfish}, Topical-Chat~\cite{gopalakrishnan:interspeech2019:topical}, and DailyDialog datasets~\cite{li:ijcnlp2017:dailydialog}, which we consider not to bring any ethical concern. 
We added to these datasets and released a set of RGM-generated responses along with binary labels denoting the presence or absence of contradictions.
Regarding the responses, it is conceivable that the aggressive expressions generated by the RGMs may be present in the collected contradictory and noncontradictory responses.
As for the labels, we have meticulously removed any personal information belonging to the workers to share our dataset ethically.

\section*{Limitations}
Our compiled dataset exclusively comprises contradictions in RGM responses to dialogue contexts that conclude with FQs extracted from the MSC dataset.
The experiments detailed in Section~\ref{sec:experiments} demonstrated that utilizing our dataset addressed RGM contradictions effectively, even for the responses to non-FQ dialogue contexts sourced from different corpora.
However, broadening the collection of contradictory responses to encompass a diverse array of dialogue contexts could facilitate a more comprehensive analysis of RGM contradictions and potentially lead to enhanced contradiction suppression through a data-driven approach.

Furthermore, it remains unclear whether employing our gathered dataset can effectively mitigate contradictory responses by any RGM.
Our data compilation involved the responses from various recent and representative high-performance RGMs.
More importantly, Section~\ref{sec:experiments} outlined that the contradiction detector, trained using our dataset, effectively identified contradictory responses from unknown RGMs.
Nonetheless, we must acknowledge the possibility that it might struggle to suppress contradictions in responses from newer RGMs.
Despite this uncertainty, we believe collecting contradictory responses from recent high-performance RGMs is crucial for developing dialogue systems that can generate consistent responses.

\section*{Acknowledgments}
This work was partly supported by JSPS KAKENHI Grant Numbers JP22K17943, JP21J22383, and
JST Moonshot R\&D Grant Number JPMJMS2011-35 (fundamental research).

\bibliography{references}

\clearpage
\appendix

\section{Contradiction types}
\label{appendix:contradiction-types}
Two major contradiction types are identified in the context of dialogue response generation: (i) contradictions against the facts in the world outside of the ongoing dialogue (e.g., personas) and (ii) those against what is stated in the local preceding context (e.g., opinions)~\cite{li:acl2020:dontsaythat,nie:acl2020:i-like-fish}.
This study focuses on suppressing the second type.
While several studies have addressed the issue of avoiding the first contradiction type~\cite{li:acl2016:personabasedneuralconv, zhang:acl2018:personachat, qian:ijcai2018:assign-persona-profile, kottur:ijcai2017:explorepersonaconv, kim:emnlp2020:will-i-sound-like-me}, given that the multi-turn human-bot interaction is attracting increasing interest, we believe that tackling the issue of the second type is becoming increasingly important.


\section{FQ analysis in existing dataset}
\label{appendix:whyfq}

We randomly examined $50$ responses from a pool of $382$ RGM-generated contradictory responses in the human-bot dialogues collected by \citet{nie:acl2020:i-like-fish}.
Remarkably, $25$ ($50\%$) of these $50$ contradictory responses were elicited by FQs, strongly indicating that FQ plays a prominent role in provoking RGM contradictions.


\section{Preliminary experiment}
\label{appendix:validate-fq}

\subsection{Experimental procedures}
We first extract samples from a pool of $C$ for three cases (i.e., $d_{u_r}=1$, $3$, and $5$) by random sampling (\texttt{RANDOM}) and picking samples with the highest FQness (\texttt{TOP}).
Subsequently, we employ multiple RGMs to generate responses to the $C$ collected by \texttt{RANDOM} and \texttt{TOP}.
We then compare the number of the RGM responses contradicting the $C$ obtained through the two abovementioned methods.

\subsection{Experimental settings}
\paragraph{Settings of dialogue context preparation.}
We utilized the pool of $C$ described in Section~\ref{subsubsec:settings-inputs} as the source for extracting instances using \texttt{RANDOM} and \texttt{TOP}.
Each \texttt{TOP} and \texttt{RANDOM} extracted $100$, $100$, and $50$ samples from the pool for $d_{u_r}=1$, $3$, and $5$, respectively.

\paragraph{Settings for RGM response collection.}
Seven RGMs were employed to generate the responses for each $C$: Plato-2, Plato-XL, Blenderbot1-3B, Blenderbot2-3B, Blenderbot3-3B, Blenderbot3-30B, and Opt-66B.\footnote{Note that the ChatGPT API service was not yet available when the preliminary experiment was conducted.}
We specifically had each RGM generate $100$ response candidates for each input through top-p sampling~\cite{holtzman:iclr2020:nucleus}, with a value of p set to $0.5$.
We chose the response with the highest generation probability among the $100$ candidates.

\paragraph{Settings for RGM response annotation.}
Each RGM response was manually assessed to determine its consistency with the context.
Amazon Mechanical Turk's $10$ workers assigned to each response performed a binary classification task to distinguish between the contradictory and noncontradictory responses.
We solely focused on evaluating the consistency with $u_r$ due to cost considerations.

\subsection{Experimental results}
Table~\ref{table:fq_effectiveness} displays the comparison results, which confirmed that more contradiction labels were assigned to the responses for $C$ with a high FQness.
This observation underscored the tendency of $C$ with a higher FQness to provoke more RGM contradictions.

\begin{table}[t]
    \small
    \centering
    \tabcolsep 6pt
    \begin{tabular}{lrrrr}
    \toprule
         & $T=1$ & $T=2$ &
           $T=3$ & $T=4$\\ 
    \midrule
        \texttt{RANDOM} & 194\,/\,700       & \ph{0}67\,/\,700  & \ph{0}33\,/\,700       & \ph{0}11\,/\,700 \\
                        & (27.7\%)          & (9.6\%)           & (4.7\%)                & (1.6\%) \\
        \texttt{TOP}    & \tb{238}\,/\,700 & \tb{101}\,/\,700 & \ph{0}\tb{50}\,/\,700 & \ph{0}\tb{23}\,/\,700 \\
                        & (\tb{34.0}\%)    & (\tb{14.4}\%)    & (\tb{7.1}\%)          & (\tb{3.3}\%) \\
    \bottomrule
    \multicolumn{5}{c}{(a) $d_{u_r}=1$}
    \end{tabular}
    \vspace*{0.05cm}
    \begin{tabular}{lrrrr}
    \toprule
         & $T=1$ & $T=2$ &
           $T=3$ & $T=4$\\ 
    \midrule
        \texttt{RANDOM} & 246\,/\,700       & \ph{0}77\,/\,700  & \ph{0}31\,/\,700       & \ph{0}10\,/\,700 \\
                        & (35.1\%)          & (11.0\%)           & (4.4\%)                & (1.4\%) \\  
        \texttt{TOP}    & \tb{270}\,/\,700 & \tb{141}\,/\,700 & \ph{0}\tb{81}\,/\,700 & \ph{0}\tb{43}\,/\,700 \\
                        & (\tb{38.6}\%)    & (\tb{20.1}\%)     & (\tb{11.6}\%)         & (\tb{6.1}\%) \\
    \bottomrule
    \multicolumn{5}{c}{(b) $d_{u_r}=3$}
    \end{tabular}
    \vspace*{0.05cm}
    \begin{tabular}{lrrrr}
    \toprule
         & $T=1$ & $T=2$ &
           $T=3$ & $T=4$\\ 
    \midrule
        \texttt{RANDOM} & \ph{0}98\,/\,350  & \ph{0}20\,/\,350       & \ph{00}6\,/\,350          & \ph{00}3\,/\,350 \\
                        & (28.0\%)          & (5.7\%)                & (1.7\%)                   & (0.9\%) \\  
        \texttt{TOP}    & \tb{126}\,/\,350 & \ph{0}\tb{50}\,/\,350 & \ph{0}\tb{25}\,/\,350 & \ph{0}\tb{17}\,/\,350 \\
                        & (\tb{36.0}\%)    & (\tb{14.3}\%)         & (\tb{7.1}\%)             & (\tb{4.9}\%) \\
    \bottomrule
    \multicolumn{5}{c}{(c) $d_{u_r}=5$}
    \end{tabular}
    \caption{The number of contradictory responses to $C$ extracted by \texttt{RANDOM} and \texttt{TOP}. Each value denotes the count of responses judged contradictory to $u_r$ by at least $T$ workers out of $10$.}
    \label{table:fq_effectiveness}
\end{table}


\begin{table*}[t]
    \centering
    \small
    \tabcolsep 9.4pt
    \begin{tabular}{lcccccccccc}
        \toprule
          & & P2 & PX & B1 & B2 & B3 & BL & O6 & CG & Total \\
        \midrule
         $d_{u_r}=1$  & \# of C & \phantom{0}840 & \phantom{0}845 & 1263           & 1526           & 1472
                          & \phantom{0}908 & 1177           & \phantom{00}77 & \phantom{0}8108 (2703) \\
                & \# of N & 1759           & 1726           & 1172           & \phantom{0}967 & 1028
                          & 1628           & 1420           & 2771           & 12471 (2920) \\
        \midrule
         $d_{u_r}=3$  & \# of C & \phantom{0}208 & \phantom{0}301 & \phantom{0}362 & \phantom{0}395 & \phantom{0}361
                          & \phantom{0}230 & \phantom{0}287 & \phantom{00}31 & \phantom{0}2175 (\phantom{0}739) \\
                & \# of N & \phantom{0}629 & \phantom{0}505 & \phantom{0}451 & \phantom{0}402 & \phantom{0}433
                          & \phantom{0}601 & \phantom{0}524 & \phantom{0}833 & \phantom{0}4378 (\phantom{0}953) \\
        \midrule
         $d_{u_r}=5$  & \# of C & \phantom{00}26 & \phantom{00}30 & \phantom{00}42 & \phantom{00}35 & \phantom{00}27
                          & \phantom{00}22 & \phantom{00}33 & \phantom{000}5 & \phantom{00}220 (\phantom{00}74) \\
                & \# of N & \phantom{00}56 & \phantom{00}47 & \phantom{00}42 & \phantom{00}44 & \phantom{00}46
                          & \phantom{00}56 & \phantom{00}50 & \phantom{00}81 & \phantom{00}422 (\phantom{00}94) \\
        \midrule
         Total  & \# of C & 1074           & 1176           & 1667           & 1956           &           1860
                          & 1160           & 1497           & \phantom{0}113 & 10503 (3516) \\
                & \# of N & 2444           & 2278           & 1665           & 1413           &           1507
                          & 2285           & 1994           & 3685           & 17271 (3967) \\
        \bottomrule
    \end{tabular}
    \caption{The number of responses in our dataset. ``\# of C'' and ``\# of N'' denote the numbers of contradictory and noncontradictory responses, respectively. Values in parentheses refer to the number of types of contexts.}
    \label{tab:dataset-statistics-detailed}
\end{table*}


\section{Settings for dataset construction}
\label{appendix:generation-settings}
Each of the eight RGMs generated one response to an input, resulting in eight responses for each $C$.
We enhanced the efficiency of gathering the contradictions by choosing the final response of an RGM to an input from the top $100$ candidates with the highest contradiction probability predicted by the state-of-the-art contradiction detector~\cite{nie:acl2020:i-like-fish}.
We utilized top-p sampling to collect the $100$ candidates.
We set a value of p to $0.5$, which was lower than the default value of $0.9$ used in major platforms, such as ParlAI~\cite{miller:emnlp2017demo:ParlAI}, to avoid sampling responses with low generation probabilities.
This allowed us to gather candidates with a high generation probability by the RGM and a high likelihood of being contradictory.
We employed OpenAI's API\footnote{\url{https://platform.openai.com}.} for ChatGPT, Knover\footnote{\url{www.github.com/PaddlePaddle/Knover}.} for Plato-2 and Plato-XL, and ParlAI for the others.

\section{Worker selection for dataset construction}
\label{appendix:worker-selection}
We first presented a task with obviously correct answers.
It contained $21$ dialogue responses requiring classification into contradictory or noncontradictory according to their preceding referent $r$.
We exclusively handpicked workers who scored fewer than two incorrect answers in this task.


\section{Details of collected dataset}
\label{appendix:breakdown}
Table~\ref{tab:dataset-statistics-detailed} illustrates the number of contradictory and noncontradictory responses obtained from each of the eight RGMs outlined in Section~\ref{sec:data-collection}.


\section{Frequency of intra-utterance inconsistencies}
\label{appendix:intra}

\begin{table*}[t]
    \centering
    \small
    \tabcolsep 10.8pt
    \begin{tabular}{lccccccccc}
        \toprule
          RGM & P2 & PX & B1 & B2 & B3 & BL & O6 & CG & Human \\
          \midrule
          Frequency & 4\,/\,50 & 5\,/\,50 & 6\,/\,50 & 8\,/\,50 & 12\,/\,50 & 8\,/\,50 & 5\,/\,50 & 0\,/\,50 & 0\,/\,200 \\
          & (8\%) & (10\%) & (12\%) & (16\%) & (24\%) & (16\%) & (10\%) & (0\%) & (0\%) \\
        \bottomrule
    \end{tabular}
    \caption{The frequency of intra-utterance inconsistencies in contradictory responses of RGMs and humans. The column labeled ``Human'' represents the frequency of intra-utterance inconsistencies in human-written contradictory responses extracted from the DECODE dataset.}
    \label{tab:count-intra}
\end{table*}

Table~\ref{tab:count-intra} illustrates the frequency of intra-utterance inconsistencies in $50$ randomly sampled contradictory responses of each of the eight RGM in our dataset and $200$ randomly extracted human-written contradictory responses from the DECODE dataset.
The counting of intra-utterance inconsistencies was carried out by the author.

Our findings indicated that seven RGMs generated at least $4$ ($8\%$) contradictory responses featuring an intra-utterance inconsistency.
The sole exception was ChatGPT (CG), whose set of contradictory responses did not encompass intra-utterance inconsistency.
This suggests that a small number of large-scale RGMs, such as ChatGPT, are progressing toward eradicating inconsistencies within individual utterances.
Nevertheless, even a sophisticated model like Opt-66B generates contradictions with intra-utterance inconsistency.

Conversely, none of the $200$ responses randomly sampled from the DECODE dataset exhibited an intra-utterance inconsistency (``Human'' in Table~\ref{tab:count-intra}).
These results suggest contradictory responses featuring intra-utterance inconsistencies are particularly frequent in RGM responses.


\begin{table*}[t]
    \small
    \centering
    \tabcolsep 4.4mm
    \begin{tabular}{lccccc}
    \toprule
        & \multicolumn{5}{c}{\footnotesize{\# of randomly sampled contradictory / noncontradictory instances for training}}    \\ \cline{2-6} \\[-1.8ex]
        Detector    & SNLI & MultiNLI & DialogueNLI & AdversarialNLI-R3 & DECODE \\ \midrule
    $\mathrm{CD_{SNLI+MNLI}}$ & 4012 / 4012   & 4011 / 4011       & \ph{000}0 / \ph{000}0                    & \ph{000}0 / \ph{000}0                         & \ph{000}0 / \ph{000}0            \\
    $\mathrm{CD_{ALL}}$       & 1004 / 1004   & 1003 / 1003       & 2006 / 2006            & 2005 / 2005                   & 2005 / 2005      \\
    $\mathrm{CD_{ALL-DNLI}}$  & 1338 / 1338   & 1337 / 1337       & \ph{000}0 / \ph{000}0                    & 2674 / 2674                   & 2674 / 2674      \\
    $\mathrm{CD_{ALL-ANLI}}$  & 1338 / 1338   & 1337 / 1337       & 2674 / 2674            & \ph{000}0 / \ph{000}0                          & 2674 / 2674      \\
    $\mathrm{CD_{ALL-DEC}}$   & 1338 / 1338   & 1337 / 1337       & 2674 / 2674            & 2674 / 2674                   & \ph{000}0 / \ph{000}0             \\
    $\mathrm{CD_{DNLI}}$      & \ph{000}0 / \ph{000}0          & \ph{000}0 / \ph{000}0              & 8023 / 8023            & \ph{000}0 / \ph{000}0                          & \ph{000}0 / \ph{000}0             \\
    $\mathrm{CD_{ANLI}}$      & \ph{000}0 / \ph{000}0          & \ph{000}0 / \ph{000}0              & \ph{000}0 / \ph{000}0                   & 8023 / 8023                   & \ph{000}0 / \ph{000}0             \\
\bottomrule
    \end{tabular}
    \caption{The number of contradictory and noncontradictory instances gathered for the training of each baseline detector. Following the experiment by \citet{nie:acl2020:i-like-fish}, training instances were randomly sampled from five datasets: the SNLI~\cite{bowman:emnlp2015:snli}, MultiNLI~\cite{williams:naacl2018:mnli}, DialogueNLI~\cite{welleck:acl2019:dialog-NLI}, AdversarialNLI-R3~\cite{nie:acl2020:anli}, and DECODE dataset. In gathering negative (contradictory) instances from Natural Language Inference (NLI) datasets, the premise text of an NLI instance labeled as contradiction was designated as $u_r$, and the hypothesis text of the same NLI instance was considered the corresponding contradictory response. Similarly, in the extraction of positive (noncontradictory) instances from NLI datasets, entailment- or neutral-labeled NLI instances were utilized. In this case, the premise text of the NLI instance was treated as $u_r$, and the hypothesis text was regarded as the noncontradictory response.}
    \label{tab:train-breakdown}
\end{table*}

\section{Settings of dialogue act labeling}
\label{appendix:dialogue-act}

For the analysis in Section~\ref{subsec:analyze-context}, we developed a labeler to assign dialogue act labels to the utterances $u_q$ in our dataset.

\subsection{Development of dialogue act labeler}

\paragraph{Inputs and outputs.}
Suppose that the $t$-th utterance $u_t$ in a dialogue consists of the $n$ segments $(u_{t,1},u_{t,2},...,u_{t,n})$ and that each segment is assigned one dialogue act label.
In addition, let the utterance immediately preceding $u_t$ be $u_{t-1}$.
We designed a labeler to predict the dialogue act label of the $i$-th segment $u_{t,i}$ within $u_t$, using $u_{t-1}$ and the segments of $u_t$ up to the $i$-th segment $(u_{t,1},u_{t,2},...,u_{t,i})$ as inputs.

\paragraph{Dataset for training.}
We developed a labeler by fine-tuning RoBERTa.\footnote{\url{https://huggingface.co/FacebookAI/roberta-large}.}
We employed the Switchboard Dialogue Act Corpus with the $42$ clustered SWBD-DAMSL dialogue act labels~\cite{jurafsky:97:switchboard} for fine-tuning.
The preprocessed dataset we used for this study\footnote{\url{https://github.com/shreyangshu12/Dialogue-act-classification}.} comprises training data with $192,\!390$ segments, validation data with $3,\!272$ segments, and evaluation data with $4,\!078$ segments.

\paragraph{Hyperparameters.}
We fine-tuned RoBERTa by employing the implementation of Hugging Face~\cite{debruyn:coling2022:huggingface} with its default settings, excluding a few parameters.\footnote{ \texttt{early\_stopping\_patience:}\,$2$, \texttt{learning\_rate:}\,$1$e-$5$, \texttt{train\_batch\_size:}\,$256$, and \texttt{weight\_decay:}\,$0.01$}

\paragraph{Performance of developed labeler.}
The test set accuracy on the aforementioned preprocessed dataset, used for fine-tuning RoBERTa, was $80.4\%$.

\subsection{Dialogue act labeling for our dataset}
The developed labeler was utilized to assign dialogue act labels to the utterance $u_q$ in our dataset.
To simplify the process, we treated each sentence in $u_q$ as a segment,%
\footnote{We split an utterance into sentences using NLTK sentence tokenizer~\cite{bird:acl2004:nltk}.}
assigning a dialogue act label to each sentence.
We used the utterance immediately preceding $u_q$ and all sentences up to the $i$-th in $u_q$ as input in order to predict the dialogue act label for the $i$-th sentence in $u_q$.
This process was applied to all sentences within $u_q$.
In our analysis in Section~\ref{subsec:analyze-context}, for the sake of simplicity, we regarded that a certain label had been assigned to $u_q$ if one or more sentences within $u_q$ were assigned that specific label.


\section{Settings for test set construction}
\label{appendix:testset}
We first constructed two pools of $C$ by extracting $d_{u_r}=1,3$ or more consecutive utterances from the Topical-Chat dataset and the DailyDialog dataset, respectively, ensuring that the final utterance contains questions.
From each of the two pools, we randomly sampled those consecutive utterances.
Specifically, for the Topical-Chat dataset, we sampled $300$ and $100$ samples from the pool for $d_{u_r}=1$ and $3$, respectively.
Similarly,  for the DailyDialog dataset, we sampled $200$ and $100$ samples from the pool for $d_{u_r}=1$ and $3$, respectively.
The other settings are the same as in our large-scale dataset construction described in Section~\ref{sec:data-collection}, except for the method of collecting $C$ described above and that responses of ChatGPT were not collected.


\section{Settings of detector training}
\label{appendix:training}

\paragraph{Samples for training.}
A negative pair comprised a contradictory response in our dataset and the preceding utterance $u_r$.
In contrast, a positive pair comprised a noncontradictory response from our dataset and one randomly selected from its preceding utterances by the same speaker.
This was because the responses annotated as noncontradictory with $u_r$ were also likely to be noncontradictory with the other preceding statements.
By introducing randomness into the selection of preceding utterances for pairing with a noncontradictory RGM response, we aimed to create positive pairs comprising unrelated utterances.
These pairs could be valuable for training detectors to recognize that unrelated pairs should be categorized as noncontradictory.

\begin{table*}[t]
    \centering
    \small
    \tabcolsep 2.58mm
    \begin{tabular}{lcclccccccccc}
        \cmidrule[0.9pt]{1-2} \cmidrule[0.9pt]{4-12}
            Detector            & By-Human  &           & Detector &P2        & PX        & B1
                                & B2        & B3        & BL        & O6        & CG        \\
        \cmidrule{1-2} \cmidrule{4-12}
            $\mathrm{CD_{SNLI+MNLI}}$ & .777 & & $\mathrm{CD_{SNLI+MNLI}}$ & .675 & .600 & .675 & .615 & .620 & .630 & .575 & .650 \\
            $\mathrm{CD_{ALL}}$       & .935 & & $\mathrm{CD_{ALL}}$       & .705 & .595 & .615 & .550 & .675 & .645 & .625 & .730 \\
            $\mathrm{CD_{ALL-DNLI}}$  & .930 & & $\mathrm{CD_{ALL-DNLI}}$  & .690 & .615 & .650 & .640 & .635 & .615 & .635 & .690 \\
            $\mathrm{CD_{ALL-ANLI}}$  & .934 & & $\mathrm{CD_{ALL-ANLI}}$  & .705 & .645 & .620 & .550 & .680 & .605 & .640 & .650 \\
            $\mathrm{CD_{ALL-DEC}}$   & .856 & & $\mathrm{CD_{ALL-DEC}}$   & .690 & .625 & .635 & .630 & .625 & .600 & .655 & .720 \\
            $\mathrm{CD_{DNLI}}$      & .771 & & $\mathrm{CD_{DNLI}}$      & .595 & .580 & .575 & .570 & .600 & .540 & .550 & .565 \\
            $\mathrm{CD_{ANLI}}$      & .790 & & $\mathrm{CD_{ANLI}}$      & .680 & .595 & .635 & .605 & .595 & .600 & .610 & .615 \\
            $\mathrm{CD_{DEC}}$       & .952 & & $\mathrm{CD_{DEC}}$       & .600 & .575 & .615 & .540 & .655 & .555 &.565  & .650 \\
            $\mathrm{CD_{OUR}}$       & .842 & & $\mathrm{CD_{OUR}}$       &\tb{.800} &\tb{.735} &.\tb{715}
                                      &\tb{.750} &\tb{.765} &\tb{.745} &\tb{.690} &\tb{.790} \\
        \cmidrule{1-2} \cmidrule{4-12}
            $\mathrm{CD_{DECfull}}$ &\tb{.958}&           & $\mathrm{CD_{DECfull}}$&.625       &.620       &.565
                                &.585       &.595       &.625       &.575       &.715       \\
        \cmidrule[0.9pt]{1-2} \cmidrule[0.9pt]{4-12}
        \multicolumn{2}{c}{(a) Human-written test set.} & &
        \multicolumn{9}{c}{(b) In-domain test sets. Scores for each target RGM are presented.} \\
    \end{tabular}
    \tabcolsep 2.53mm
    \begin{tabular}{lccccccccc}
        \toprule
                                & \footnotesize{Test set from Nie+'21}  &
                                & \multicolumn{7}{c}{\footnotesize{Test sets from Topical-Chat\,/\,DailyDialog}} \\
        \cline{2-2} \cline{4-10} \\[-1.8ex]
            Detector            & Human-Bot &           & P2                 & PX                 & B1
                                & B2                    & B3                 & BL                 & O6 \\
        \midrule
            $\mathrm{CD_{SNLI+MNLI}}$ & .729 & & .58/.66 & .63/.60 & .59/.54 & .53/.64 & .58/.64 & .60/.58 & .61/.67 \\
            $\mathrm{CD_{ALL}}$       & .783 & & .63/.66 & .63/.64 & .67/.54 & .61/.53 & .63/.62 & .72/.56 & .61/.59 \\
            $\mathrm{CD_{ALL-DNLI}}$  & .783 & & .65/.69 & .65/.62 & .69/.57 & .61/.56 & .71/.69 & .66/.56 & .55/.65 \\
            $\mathrm{CD_{ALL-ANLI}}$  & .775 & & .68/.66 & .61/.64 & .61/.61 & .60/.63 & .70/.66 & .70/.56 & .60/.57 \\
            $\mathrm{CD_{ALL-DEC}}$   & .771 & & .60/.67 & .57/.63 & .56/.55 & .57/.65 & .64/.66 & .72/.62 & .62/.62 \\
            $\mathrm{CD_{DNLI}}$      & .736 & & .60/.60 & .62/.56 & .57/.60 & .60/.62 & .63/.53 & .62/.54 & .71/.55 \\
            $\mathrm{CD_{ANLI}}$      & .688 & & .63/.64 & .70/.69 & .65/.57 & .61/.69 & .68/.62 & .65/.61 & .59/.66 \\
            $\mathrm{CD_{DEC}}$       & .749 & & .55/.52 & .58/.60 & .61/.55 & .60/.59 & .68/.61 & .67/.55 & .59/.53 \\
            $\mathrm{CD_{OUR}}$       & .787 & &\tb{.77}/\tb{.77} &\tb{.75}/\tb{.71} &\tb{.74}/\tb{.68}
                                &\tb{.70}/\tb{.72}    &\tb{.73}/\tb{.76} &\tb{.82}/\tb{.64} &\tb{.81}/\tb{.75} \\
        \midrule
            $\mathrm{CD_{DECfull}}$ & \tb{.829}      &           &.57/.52             &.64/.59             &.67/.56
                                &.64/.59                &.69/.66             &.71/.54             &.59/.54 \\
        \bottomrule
        \multicolumn{10}{c}{(c) Out-of-domain test sets. For the Topical-Chat and DailyDialog test sets, scores for each target RGM are presented.} \\
    \end{tabular}
    \caption{Accuracy of $\mathrm{CD_{OUR}}$ and all rival detectors for (a) human-written, (b) in-domain, and (c) out-of-domain test sets.}
    \label{table:detector-accuracy-indomain-full}
\end{table*}

\paragraph{Hyperparameters.}
We fine-tuned RoBERTa\footnote{\url{https://huggingface.co/FacebookAI/roberta-large}.} by employing the implementation of Hugging Face~\cite{debruyn:coling2022:huggingface} with its default settings, excluding a few parameters.\footnote{\texttt{train\_batch\_size:}\,$128$, \texttt{weight\_decay:}\,$0.01$, \texttt{eval\_steps:}\,$200$, \texttt{early\_stopping\_patience:}\,$1$, and \texttt{learning\_rate:}\,\{$1$e-$6$, $5$e-$6$, $1$e-$5$, $5$e-$5$\}.}
We updated the model parameters until we reached a point where early stopping was triggered.
Early stopping was determined by assessing the accuracy of validation data, a distinct subset comprising $10\%$ of the training data and withheld from the training process.
We saved the model parameters with the highest accuracy on the validation data at each learning rate and ultimately selected that with the highest validation accuracy among all the saved parameters.

\section{Comparison to diverse rival detectors}
\label{appendix:additional-experiments}

Table~\ref{table:detector-accuracy-indomain-full} presents the outcomes of evaluating the contradiction detection capabilities of $\mathrm{CD_{OUR}}$ in comparison to diverse rival detectors.

\paragraph{Comparison to baselines employed in \citet{nie:acl2020:i-like-fish}.}
In addition to $\mathrm{CD_{DEC}}$, we developed seven detectors named $\mathrm{CD_{SNLI+MNLI}}$, $\mathrm{CD_{ALL}}$, $\mathrm{CD_{ALL-DNLI}}$, $\mathrm{CD_{ALL-ANLI}}$, $\mathrm{CD_{ALL-DEC}}$, $\mathrm{CD_{DNLI}}$, and $\mathrm{CD_{ANLI}}$, corresponding to the seven RoBERTa-based baseline detectors utilized in \citet{nie:acl2020:i-like-fish}'s experiments.
The training settings for all detectors were consistent with those of $\mathrm{CD_{DEC}}$, with the only variation being the source of the training data (Table~\ref{tab:train-breakdown}).
We subsequently juxtaposed these detectors with $\mathrm{CD_{OUR}}$.
Table~\ref{table:detector-accuracy-indomain-full} illustrates that $\mathrm{CD_{OUR}}$, formed using our dataset, exhibited a higher accuracy in identifying RGM contradictory responses compared to those rival detectors.

\paragraph{Comparison to a detector trained with more human-written data.}
Additionally, we developed the detector $\mathrm{CD_{DECfull}}$ through training on all instances within the DECODE dataset, encompassing $15,\!605$ contradictory and $15,\!605$ noncontradictory responses.
This training process followed the methodology employed for $\mathrm{CD_{DEC}}$.
Noteworthy is the observation that, despite $\mathrm{CD_{OUR}}$'s training dataset being approximately half the size of $\mathrm{CD_{DECfull}}$, it demonstrated superior performance across all RGM-generated test sets, with the exception of the Human-Bot set.
The Human-Bot test data comprises the utterances in first-meeting dialogues.
Given that $\mathrm{CD_{DECfull}}$'s training dataset also encompasses human-written contradictory responses following first-meeting dialogues, it is conceivable that the overlap in domains enabled the detector to recognize contradictions in the Human-Bot test data.

\end{document}